\documentclass[conference]{IEEEtran}
\IEEEoverridecommandlockouts
\usepackage{spconf,amsmath,graphicx,amsfonts,mathtools}
\usepackage{xcolor}
\usepackage{cite}
\usepackage{amsmath,amssymb,amsfonts}
\usepackage{algorithmic}
\usepackage{graphicx}
\usepackage{textcomp}
\usepackage{xcolor}
\usepackage{titlesec}
\def\BibTeX{{\rm B\kern-.05em{\sc i\kern-.025em b}\kern-.08em
    T\kern-.1667em\lower.7ex\hbox{E}\kern-.125emX}}

\title{A Practical Adversarial Attack on Contingency Detection of Smart Energy Systems}
\name{Moein Sabounchi and Jin Wei-Kocsis}
\address{Department of Computer and Information Technology - Purdue University}
\begin{document}
%
\maketitle
\begin{abstract}
Due to the advances in computing and sensing, deep learning (DL) has widely been applied in smart energy systems (SESs). These DL-based solutions have proved their potentials in improving the effectiveness and adaptiveness of the control systems. However, in recent years, increasing evidence shows that DL techniques can be manipulated by adversarial attacks with carefully-crafted perturbations. Adversarial attacks have been studied in computer vision and natural language processing. However, there is very limited work focusing on the adversarial attack deployment and mitigation in energy systems. In this regard, to better prepare the SESs against potential adversarial attacks, we propose an innovative adversarial attack model that can practically compromise dynamical controls of energy system. We also optimize the deployment of the proposed adversarial attack model by employing deep reinforcement learning (RL) techniques. In this paper, we present our first-stage work in this direction. In simulation section, we evaluate the performance of our proposed adversarial attack model using standard IEEE 9-bus system.
\end{abstract}
\section{Introduction}\label{sec:intro}
Due to advances in computing and sensing, DL techniques have widely been applied in control management in SESs \cite{7926429}, which has brought upon the advent of emerging tools to effectively address the uncertainties, disturbances, and unforeseen circumstances that may lead to cascading failures of the systems. However, in recent years, increasing evidence shows that DL techniques can be manipulated by adversarial attacks \cite{huang2017adversarial}. Adversarial attacks have been introduced on a variety of formats including evasion-attack \cite{biggio2013evasion}, poisoning attack \cite{quiring2020backdooring}, and inference attack \cite{shokri2017membership}. Based on the knowledge of the attackers, adversarial attacks can also be grouped into white-box, grey-box, and black-box \cite{DBLP:journals/corr/abs-1802-06806}. In recent years, adversarial attacks have been widely studied in computer vision and natural language processing \cite{akhtar2018threat,morris-etal-2020-textattack}, including TextAttack, Fast Gradient Sign Attack (FGSM), projected gradient descent (PGD), and DeepFool attacks \cite{morris-etal-2020-textattack,madry2019deep,moosavi2016deepfool,DBLP:journals/corr/CarliniW16a}. However, there is very limited work formulating adversarial attack models deployment and mitigation in SESs. As far as we know, the work in \cite{li2020conaml} is the only published work in formulating adversarial attack models in SESs. In this work, an effective constrained adversarial attack, called ConAML, is implemented to compromise cyber-physical systems. However, this proposed method mainly focuses on mitigating static operations of energy system.
In this paper, we propose an innovative black-box evasion-attack model that is able to practically mislead the decision making of contingency detection in dynamic operation of SESs. To achieve this goal, we develop a practical model to generate additive procedural-noise adversarial perturbation that is used to manipulate the power system measurements and compromise the decision making of contingency detection. To optimize the deployment of generating procedural-noise adversarial perturbation, we design an adversarial RL agent \cite{sutton2018reinforcement} to determine appropriate hyper-parameters of procedural-noise perturbation and maximize the effect of the perturbation on minimizing the accuracy of contingency detection in dynamic control of SESs. Deep RL is a type of DL, where an agent learns how to optimize the actions in an environment by observing the states and seeing the results. Various deep RL techniques have been developed for different application scenarios, such as deep Q-learning \cite{hester2018deep}, advantage actor critic \cite{babaeizadeh2016reinforcement}, and deep deterministic policy gradient (DDPG) \cite{lillicrap2015continuous}. Due to the nature of our proposed adversarial RL agent that determines optimal hyper-parameters of procedural-noise adversarial perturbation from a continuous space, we decide to employ the DDPG technique.

The rest of this paper is organized as follows, Section \ref{sec:three} will introduce our proposed black-box evasion-attack for compromising contingency detection in smart energy systems. Sections III and IV will present our performance evaluation and conclusions, respectively. 

\section{Proposed Adversarial Attack Model}\label{sec:three}
In this section, we propose an adversarial attack, specifically a black-box evasion-attack, which aims to compromise the detection of contingency in dynamic control of SESs. 
\subsection{Threat Model}\label{thmodel}
We formulate the threat model of our proposed evasion-attack as follows: 1) The attacker aims to launch an evasion-attack vector together with other cyber/physical-attack vectors to carry out sophisticated and coordinated attacks; 2) The evasion attack aims to increase the opportunity of coordinated cyber/physical attack to bypass the DL-based contingency detector of the SESs; 3) Because a contingency detector normally has high security priority for dynamic control of SESs, the DL models for contingency detector should be placed with comprehensive and advanced security measures. Therefore, it is reasonable to assume the attack cannot directly access the DL model. Instead, the attack only have limited access to the inputs and outputs of the DL model via eavesdropping and inference attacks \cite{peake2005eavesdropping,rahman2018membership}. Therefore, the proposed evasion attack is assumed to be black-box; 4) The attacker generates the black-box evasion attack by ensuring certain physical principles, such as voltage stability, for realizing the evasion attack to be stealthy; 5) The attacker can access and compromise a subset of the measurement data that are used by the DL-based contingency detector for inference. This can be realized by man-in-the-middle (MITM) attack \cite{mallik2019man}. 
\subsection{Proposed Evasion Attack Model}\label{Eva}
Based on the threat model stated above, we develop an innovative evasion attack by exploiting procedural noise \cite{lagae2009procedural} and RL techniques. The overview of our proposed work is illustrated in Fig. 1.
\begin{figure*}[!htbp]
\includegraphics[width=\linewidth]{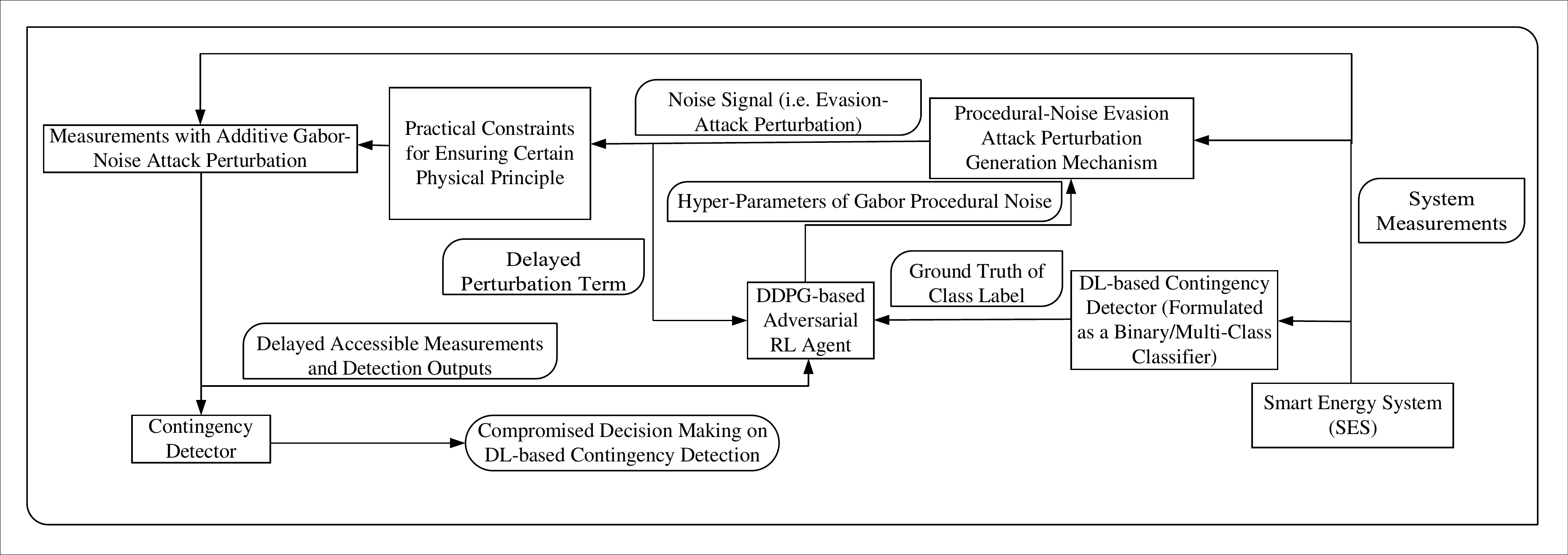}
\caption{Illustration of our proposed adversarial attack model.}
\label{fig1}
\end{figure*}
As it can be seen, the SES measurements are used to detect cyber/physical attacks or failures in the system based on a DL-based contingency detector whose behavior can be formulated as a binary/multi-class classifier \cite{hastie2009multi}. The adversarial agent on the other hand aims to mislead the decision making of the DL-based contingency detector. To achieve this goal, the attacker accesses and manipulates a subset of measurement data of the energy system by performing procedural-noise evasion-attack perturbation generation. This process is realized by exploiting Gabor procedural noise technique \cite{galerne2012gabor}. The deployment of evasion-attack perturbation generation is optimized by a DDPG-based adversarial RL agent that calculates the optimal hyper-parameters of the Gabor-noise perturbation. This adversarial RL agent generates the hyper-parameters of the procedural noise based on the ground-truth values of the measurement data that are accessible to the attacker, the currently updated Gabor-noise perturbation values, and the decision making of the contingency detector. The procedural-noise evasion-attack perturbation generation mechanism further updates the Gabor-noise evasion-attack perturbation based on the adaptively updated hyper-parameters. To realize stealthy evasion-attack, the generated Gabor-noise perturbation is further constrained based on certain physical principles of the energy system such as voltage stability. By launching the Gabor-noise evasion-attack perturbation that is adaptively optimized by the DDPG-based adversarial RL agent, the compromised measurements (i.e. measurement with additive adversarial noise) are generated to mislead the decision making on DL-based contingency detector and increase the possibility of coordinated cyber/physical attacks to bypass the detector. Therefore, as shown in Fig. 1, our proposed evasion-attack model mainly consists of two essential components: 1) procedural-noise evasion-attack perturbation generation mechanism that is designed based on Gabor procedural noise technique; and 2) DDPG-based adversarial RL agent for optimizing evasion-attack deployment. 

\subsubsection{Procedural-Noise Evasion Attack Perturbation Generation Mechanism:}
One essential challenge of practically carrying out adversarial attacks on dynamic control in energy system is that dynamic control normally requires timely response. Because of this, the gradient-based adversarial attack methods \cite{morris-etal-2020-textattack,madry2019deep,moosavi2016deepfool,DBLP:journals/corr/CarliniW16a} and the method proposed in [12] can not be directly applied in this application domain. To tackle this challenge, in our work, we exploit procedural noise model to timely generate unified evasion-attack perturbation. The main advantages of procedure noise model includes: 1) requiring low memory and computational complexity, which enables timely generation of adversarial perturbation; 2) being not dependent on a specific measurement data which enables unified adversarial-attack generation; 3) being able to be assessed and optimized independently in an online manner \cite{lagae2010survey}. Perlin \cite{perlin2002improving} and Gabor noise models are two widely adopted procedural noise models. In our work, we exploit Gabor noise model for our evasion-attack perturbation generation. 

Procedural noise is non-periodic which can be implemented in \textit{N} dimensions, where $N>1$. Procedural noise is parameterized meaning that it can generate a “class of related noise patterns” \cite{lagae2010survey}. The noise functions can generally be classified into three main groups: lattice gradient noises, explicit noises, and sparse convolution noises \cite{lagae2010survey}. For each case, we can present multiple noise functions. As stated previously, we leverage 2-D Gabor noise \cite{galerne2012gabor} in our work. Gabor Kernel is the critical component for calculating Gabor noise, which is the multiplication of a “circular Gaussian” function and two-dimensional cosine function:
\begin{equation}\label{eq1}
g(x,y)= Ke^{-\pi \sigma^2\ (x^2+y^2\ )\ } \cos\left(2\pi F_0\ (xcos\omega_0+ysin\omega_0)\right)
\end{equation}
where $K$ and $\sigma$ are the magnitude and width of the circular Gaussian function, respectively. $F_0$ and $\omega_0$ are the frequency and orientation of the cosine function, respectively. Gabor noise can be derived as the weighted sum of Gabor Kernels, which can be described as follows:
\begin{equation}\label{eq2}
N(x,y)=\sum_i W_i\ g(K_i,\sigma_i,F_{0_i},\omega_{0_i},x-x_i,y-y_i), 
\end{equation}
where $\{W_i\}$ is the set of weights and $g(\cdot)$ is the Gabor kernel. In our proposed method, we use the absolute values of the measurements in SESs, including voltage, active and reactive power, and frequency, as the x-dimension variable in Eq. (2). Additionally, we use logarithm of the bus indices of energy system as the y-dimension variable in Eq. (2), which can be formulated as follows:
\begin{equation}
y_i=\log{(i+1)}
\end{equation}
The authors would like to clarify that there can be alternative functions to $log(\cdot)$ in generating $y_i$. For simplicity, we adopt $log(\cdot)$ in our work. Additionally, as stated above, $x_i$ can be the measurements of voltage, active and reactive power, and frequency. When it is desired to generate Gabor-noise adversarial perturbation associated with more then one types of the measurements, parallel Gabor model needs to be applied. 
\subsubsection{DDPG-based Adversarial RL Agent:}\label{sec:323}
To optimize the deployment of our proposed evasion-attack, we exploit one RL technique, DDPG, to generate appropriate values of the critical hyper-parameters of the Gabor-noise evasion perturbation model, including $\sigma, F_0$, and $\omega_0$, and maximize the effect of the evasion-attack on minimizing accuracy of detecting contingency on dynamic operation of SESs. To achieve this, we model the procedural-noise perturbation generation for the energy system as a Markov decision process (MDP) that is defined by $(\mathcal{S}, \mathcal{A}, r, \mathbb{P})$. $s_k \in \mathcal{S}$ represents the input states for the system. The action $a_k \in\mathcal{A}$ refers to the decision making on the values of hyper-parameters of the Gabor-noise perturbation model including $\sigma, F_0$, and $\omega_0$ at time $t=k$. The decision making on the action results in the generation of Gabor-noise perturbation $n_k$. $\mathbb{P}$ is the transition probability that characterizes the dynamics of the SES operation. $r: \mathcal{S} \rightarrow R$ is the reward function for our proposed adversarial RL. The reward function at time $t=k$, is formulated as follows:
\begin{equation}\label{eq1}
R(k)=\sum_{\tau=k}^{T_f} \lambda^{\tau-k}R(s_\tau,a_\tau)
\end{equation}
where $s_k \in \mathcal{S}$, which as mentioned earlier represents the input states at time $t=k$, can be written per the following:
\begin{equation}\label{sk}
s_k = {\left[ x_k, n_k, c_k \right]}
\end{equation}
where $x_k$ is the vector consisting of the absolute values of the measurements associated with the individual buses (e.g. voltage, power, frequency and reactive power vectors individually) at time $t=k$. Specifically, we can describe it as: $x_{k} = \left\{x_{k,i}\right\}$, where $i$ denotes the bus index associated with the individual measurement data in the SES. $n_k$ denotes the Gabor-noise perturbation, $c_k$ denotes the difference between ground-truth and misled outputs of the contingency detector, and $\lambda$ is the discount factor. {\color{black}{ Additionally, the reward value at time $t=k$, $R(s_k,a_k)$ in Eq. (4) is formulated as follows:}}
\begin{equation}\label{reward}
R(s_k, a_k) = c_k - \sum_i e^{k_0 \left(x_{k,i}-\hat{x}\right) }-\sum_i e^{k_0 n_{k,i} } \\
\end{equation}
where $c_k, {x_{k,i}}, {n_{k,i}} \in s_k$ as defined in Eq. (\ref{sk}), and ${n_{k,i}}$ is calculated based on $a_k$, and $\hat{x}$ represents the nominal value for each measurement. In our work, we optimize the structure of the adversarial RL agent via online training, and store the trained agent model for timely usage. The proposed adversarial RL agent is then utilized to adaptively generate appropriate hyper-parameters of the Gabor-noise perturbation in near real time with minimal computational complexity.
\section{Simulation Results}\label{sec:four}
In this section, we evaluate the performance of our proposed adversarial attack, which is specifically a black-box evasion attack, on compromising the contingency detection in dynamic operation of SESs. We use IEEE 9-Bus System for the performance evaluation \cite{vittal2019power}. In the simulation, we consider that there is a built-in contingency detector based on a fully-connected deep neural network. Additionally, we assume that the attacker is able to access and manipulate the voltage measurements, i.e. $x_k$ refers to voltage measurements in our simulations. Furthermore, the attacker aims to develop and carry out the proposed Gabor-noise evasion attack to minimize the accuracy of the contingency detector on detecting a coordinated physical attack on Generator 5 beginning from $t=5.4s$. Without deploying evasion-attack, the voltage measurements are shown in Fig. 2. In this situation, the posterior probability of contingency occurrence estimated by the DL-based detector is shown in Fig. 3. We can observe that the detector is able to timely detect the fault with high accuracy when our proposed adversarial attack is not launched. 
\vspace{-4mm}
\begin{figure}[!htbp]
\includegraphics[width=\linewidth]{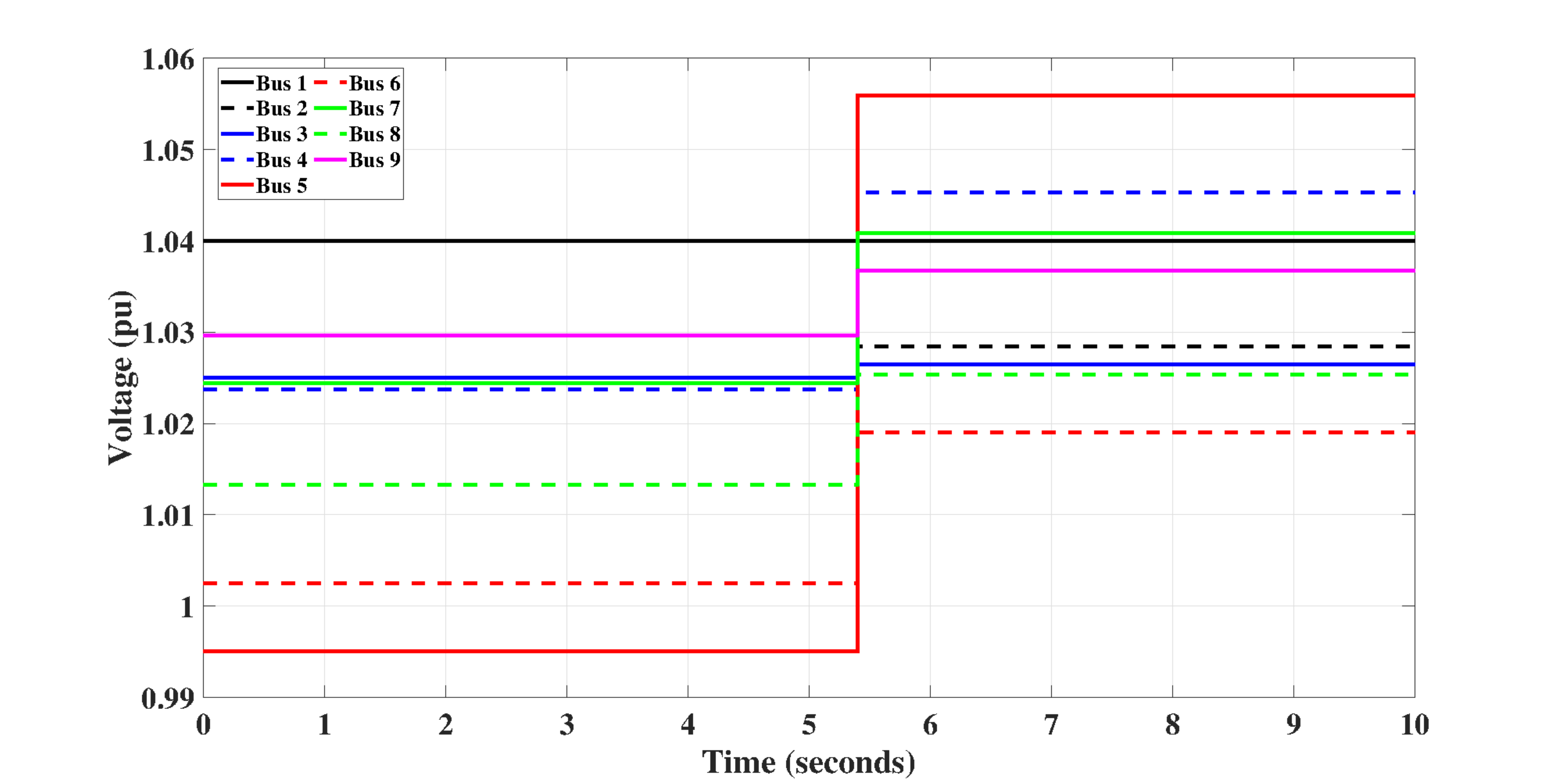}
\vspace{-8mm}
\caption{Voltage measurements with physical fault and without evasion attack.}
\label{fig2}
\end{figure}
\vspace{-3mm}
\begin{figure}[!htbp]
\includegraphics[width=\linewidth]{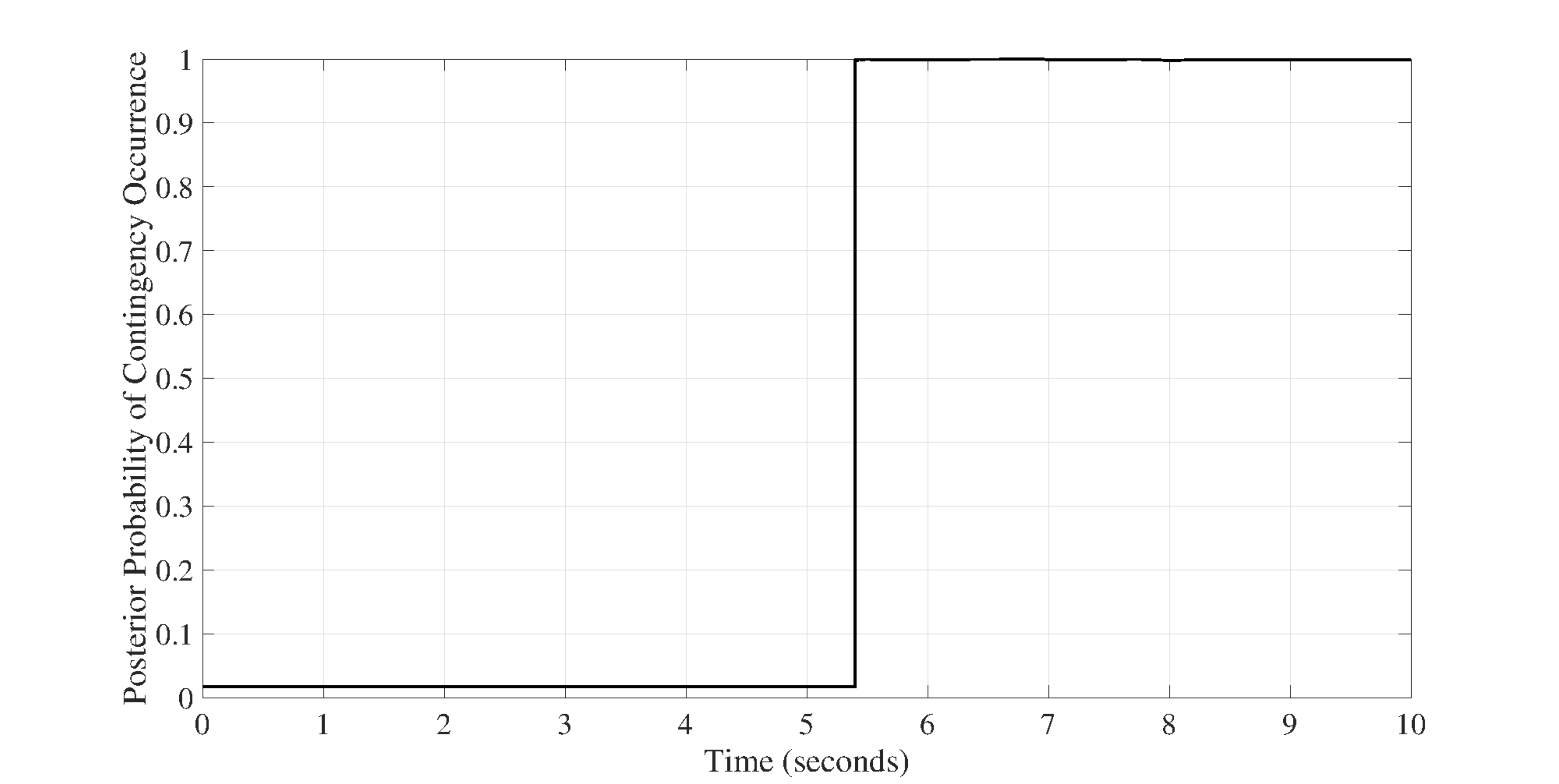}
\vspace{-8mm}
\caption{Performance of the DL-based contingency detector with physical attack.}
\label{fig3}
\end{figure}
Next, we generate our proposed Gabor-noise evasion-attack perturbation whose critical hyper-parameters are optimized by using our proposed DDPG-based adversarial RL agent. The normalized perturbation value is shown in Fig. 4. The perturbation is added to the voltage measurements shown in Fig. 2. The compromised voltage measurements are shown in Fig. 5. From Fig. 5, we can observe that the evasion-attack perturbation only results in a range of voltage-measurement variation within $[0-0.01] pu$. In other words, our proposed evasion-attack perturbation is effectively constrained based on a certain physical principle (i.e. voltage stability in our scenario). By using the compromised voltage measurements, the performance of the DL-based detector on detecting the physical attack is shown in Fig. 6. From Fig. 6, we can observe that our proposed evasion-attack can effectively mitigate the performance of the DL-based detector of the SESs. 
\vspace{-3mm}
\begin{figure}[!htbp]
\includegraphics[width=\linewidth]{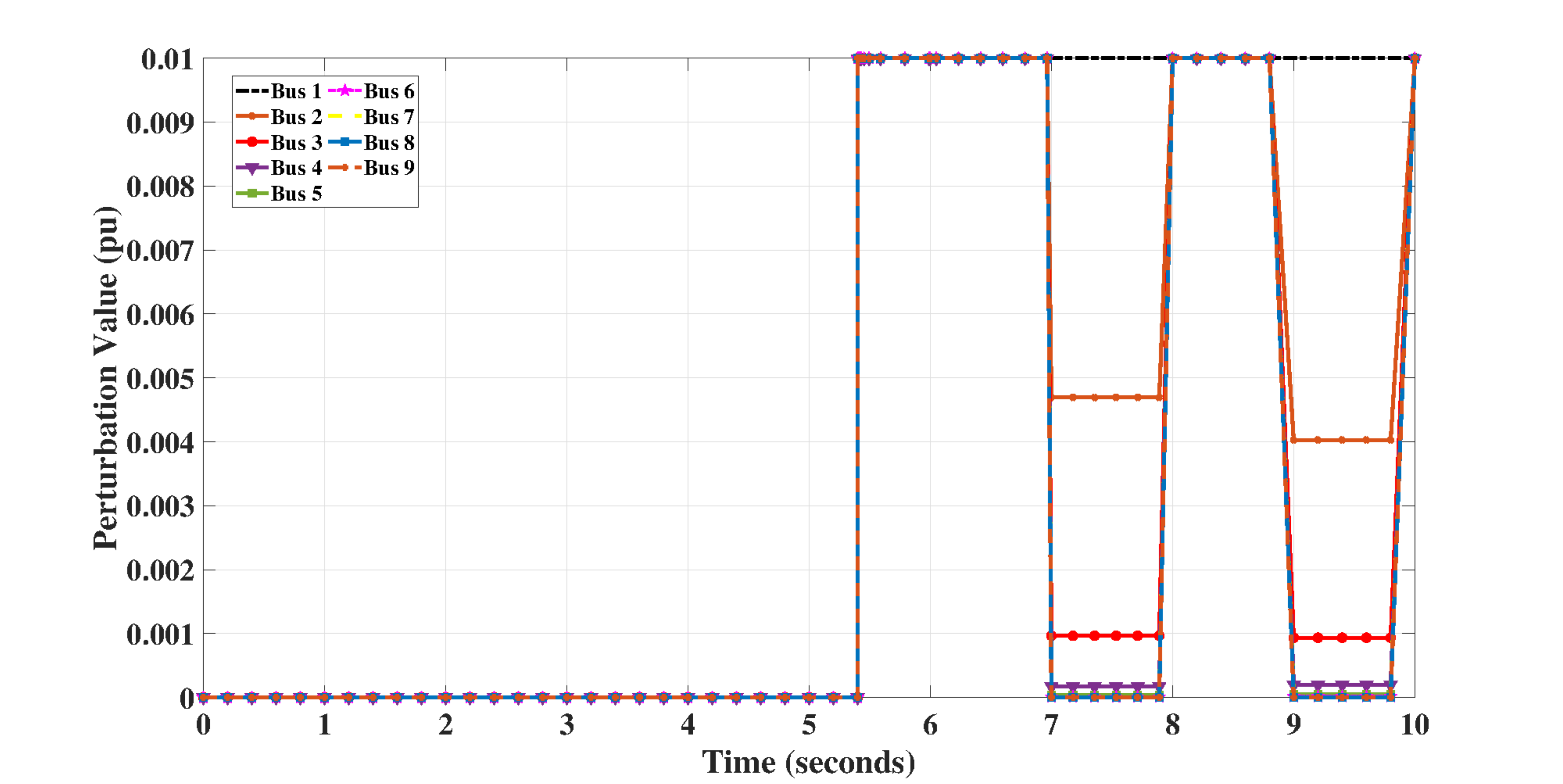}
\vspace{-8mm}
\caption{Values of Gabor-noise evasion-attack perturbation associated with individual buses.}
\label{fig4}
\end{figure}
\vspace{-6mm}
\begin{figure}[!htbp]
\includegraphics[width=\linewidth]{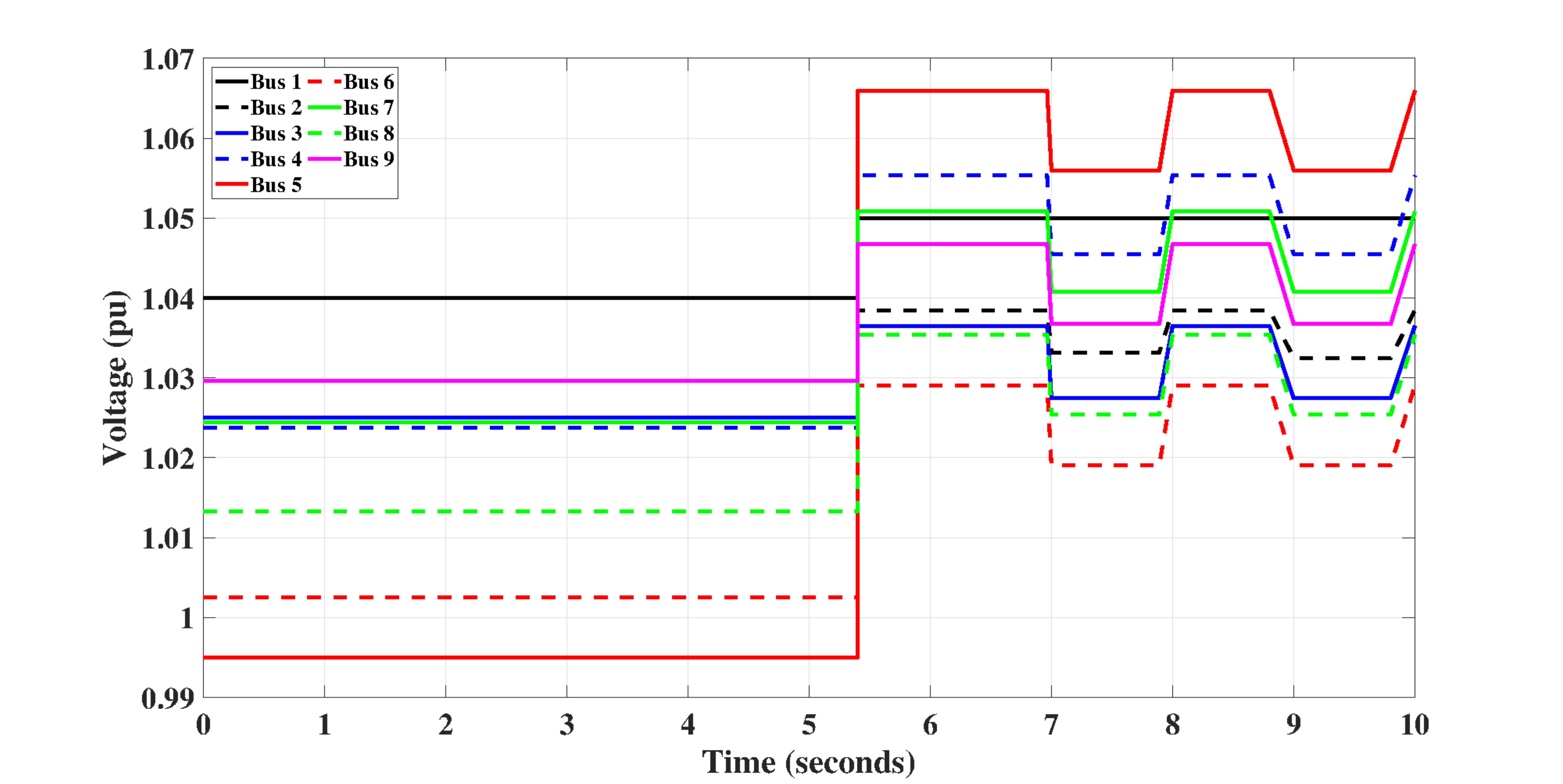}
\vspace{-8mm}
\caption{Compromised voltage measurements by launching our proposed evasion attack.}
\label{fig5}
\end{figure}
\vspace{-6mm}
\begin{figure}[!htbp]
\includegraphics[width=\linewidth]{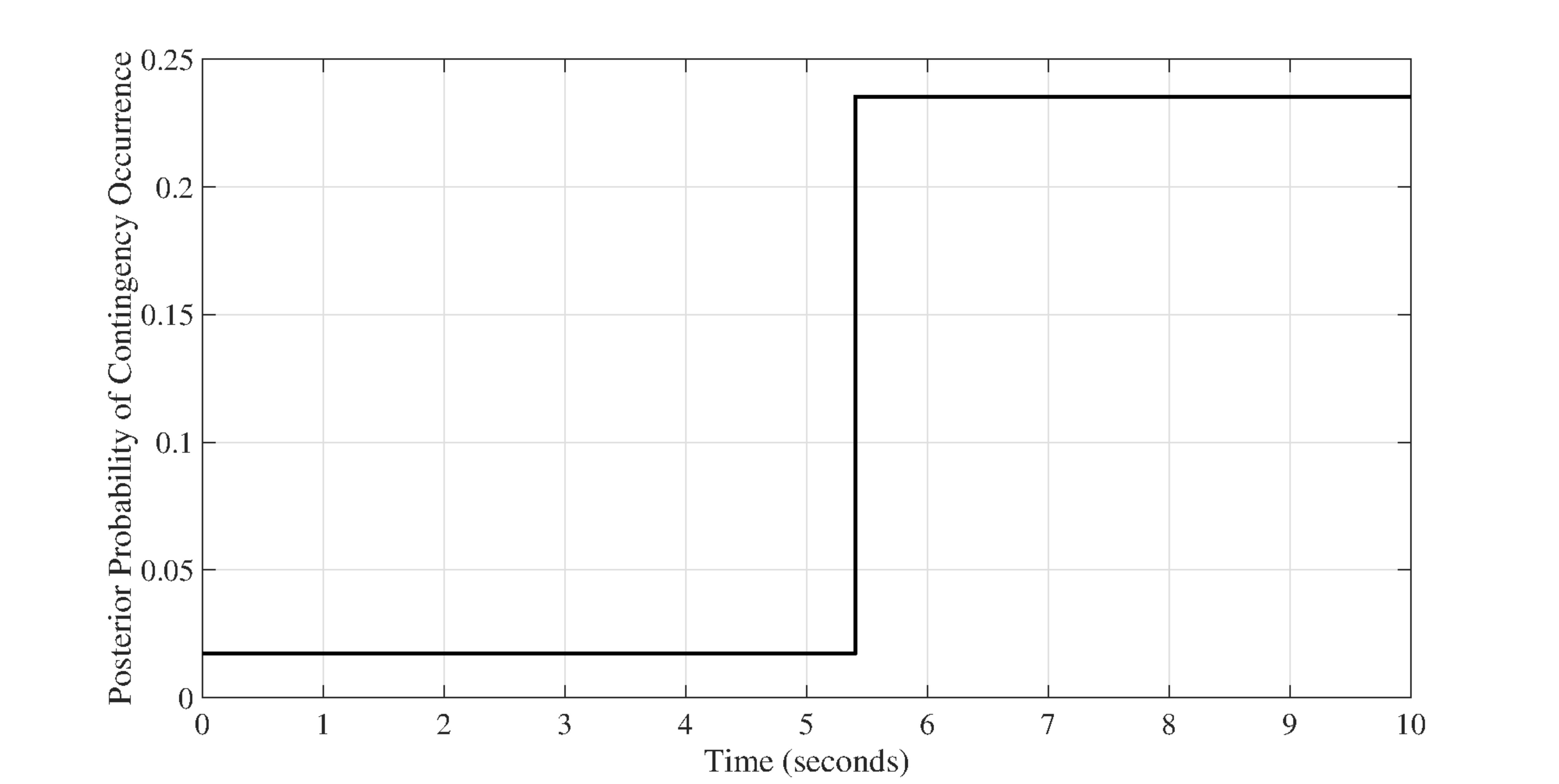}
\vspace{-8mm}
\caption{Performance of DL-based contingency detector with physical attack and evasion attack.}
\label{fig6}
\end{figure}
\section{Conclusions}\label{Conclusions}
In this paper, we present our initial-stage work on developing a practical adversarial attacks on dynamic controls of SESs. In this work, we propose an innovative adversarial attack model that is able to practically compromise the contingency detection in SESs. Our proposed model mainly consists of two components: 1) procedural-noise evasion-attack perturbation generation mechanism, and 2) DDPG-based adversarial RL agent for optimizing the evasion-attack deployment process. As illustrated in the simulation results, our proposed evasion attack model can practically compromise the contingency detector in the SESs by resulting in a significant reduction in detection accuracy. In our ongoing work, we are improving the formulation and realization of the adversarial attack by using larger-scale energy systems, and we aim to further leverage the proposed adversarial attack for enhancing the resilience of the intelligent control mechanism in energy systems.
\vspace{-1.5mm}
\bibliographystyle{IEEEbib}
\bibliography{refs}
\end{document}